\documentclass[conference]{IEEEtran}

\usepackage[colorlinks,urlcolor=blue,linkcolor=blue,citecolor=blue]{hyperref}

\usepackage{makecell, subcaption}
\usepackage{amsmath,amssymb,amsfonts}
\usepackage{pifont}% http://ctan.org/pkg/pifont
\newcommand{\cmark}{\ding{51}}%
\newcommand{\xmark}{\ding{55}}%
\usepackage{ragged2e}
\usepackage{algorithm}
\usepackage{algcompatible}

\usepackage{graphicx}
\usepackage{textcomp}
\usepackage{rotating}
\usepackage{balance}
\usepackage{multirow}
\usepackage{gensymb}
\usepackage{array,booktabs,longtable,tabularx,tabulary}
\usepackage{lipsum, balance}

\usepackage{hyphenat}

\newcommand*{\email}[1]{\normalsize\texttt{\href{mailto:#1}{#1}}\par}

\usepackage{threeparttable, booktabs, url, eqparbox, xspace, setspace, tabularx, color, soul, wrapfig}
\makeatletter
\DeclareRobustCommand\onedot{\futurelet\@let@token\@onedot}
\def\@onedot{\ifx\@let@token.\else.\null\fi\xspace}

\def\etal{\emph{et al}\onedot}

\begin{document}

\title{Emotion Recognition in Older Adults with Quantum Machine Learning and Wearable Sensors}

\thanks{--}

% uncomment before sending to blind review
%\author{\vspace{0.85in}}

% comment before sending to blind review
\author{
    \IEEEauthorblockN{%
         Md. Saif Hassan Onim$^1$,
         Travis S. Humble$^2$ and
         Himanshu Thapliyal$^1$
                    }

    \IEEEauthorblockA{%
        $^{1}$Department of Electrical Engineering and Computer Science\\
        University of Tennessee, Knoxville, TN, USA\\
        $^2$Quantum Science Center, Oak Ridge National Laboratory, Oak Ridge, TN, USA\\}
    
\email{monim@vols.utk.edu},
\email{humblets@ornl.gov},
\email{hthapliyal@utk.edu}            
    }
\maketitle

\begin{abstract}
We investigate the feasibility of inferring emotional states exclusively from physiological signals, thereby presenting a privacy-preserving alternative to conventional facial recognition techniques. We conduct a performance comparison of classical machine learning algorithms and hybrid quantum machine learning (QML) methods with a quantum kernel-based model. Our results indicate that the quantum-enhanced SVM surpasses classical counterparts in classification performance across all emotion categories, even when trained on limited datasets. The F1 scores over all classes are over 80\% with around a maximum of 36\% improvement in the recall values. The integration of wearable sensor data with quantum machine learning not only enhances accuracy and robustness but also facilitates unobtrusive emotion recognition. This methodology holds promise for populations with impaired communication abilities, such as individuals with Alzheimer’s Disease and Related Dementias (ADRD) and veterans with Post-Traumatic Stress Disorder (PTSD). The findings establish an early foundation for passive emotional monitoring in clinical and assisted living conditions.
\end{abstract}


\section{Introduction} 

Emotional well-being plays a critical role in the overall health of older adults. Recent works show substantial evidence indicating strong associations of emotional states with both cognitive and physical functioning. Deviations in emotional patterns have been identified as early markers of neurodegenerative conditions such as Alzheimer’s disease and related dementias (ADRD)~\cite{rhodus2024}. Accurate emotion detection is thus essential for the diagnosis and monitoring of psychiatric conditions, including post-traumatic stress disorder (PTSD), depression, and anxiety—particularly among older adults and other vulnerable populations. 

Traditional emotion recognition methodologies have predominantly relied on facial expression analysis, speech processing, and behavioral assessments. While these approaches have demonstrated effectiveness, they often necessitate camera-based systems or active user engagement, posing challenges related to user privacy, intrusiveness, and scalability in naturalistic settings~\cite{din2024, badawi2024}. Recent advances in wearable technologies and physiological sensing offer a compelling alternative for emotion recognition through continuous, noninvasive data collection. Physiological signals—such as electrodermal activity (EDA), heart rate variability (HRV), blood volume pulse (BVP), and skin temperature—serve as robust indicators of emotional and cognitive states. Prior research utilizing wearable devices like the Empatica E4 has demonstrated promising results in real-time detection of stress, anxiety, and cognitive load during both clinical and educational tasks~\cite{onim2023review, Delmastro2020, Kikhia2016}. In these tasks, both obtrusive and non-obtrusive instruments were used to gather physiological signals that serve as biometric indicators~\cite{onim2024utilizing, Cheong2020, Ferreira2014}.

Despite these advancements, the high dimensionality, temporal complexity, and noise inherent in physiological time-series data present substantial challenges for classical machine learning models, which are often computationally intensive and limited in their capacity to generalize across heterogeneous populations. In response to these limitations, Quantum Machine Learning (QML) has emerged as a novel computational paradigm capable of addressing the complexity of physiological data with enhanced efficiency and potentially greater expressiveness. By harnessing principles of quantum mechanics, QML models offer the potential to process high-dimensional data with reduced resources and improved representations.

\begin{table*}[htbp]
\caption{Recent Literature on Emotion Detection}
    \centering
    \resizebox{2\columnwidth}{!}{
    \setlength{\tabcolsep}{7pt}
    \begin{tabular}{clccccc}
    \toprule
        \bf Method & \makecell[c]{\bf Authors \&\\ \bf Year} & \bf Data Acquisition & \makecell[c]{\bf Detected\\ \bf Emotions} & \makecell[c]{\bf Emotion\\ \bf Stimuli} & \makecell[c]{\bf Requires Video\\ \bf for Detection ?} & \makecell[c]{\bf Target Age Group\\ \bf of Participants}\\
        
    \midrule
        \multirow{15}{*}{\rotatebox[origin=c]{90}{Classical}} & \makecell[c]{Zhang~\etal~\cite{zhang2020}\\2020} & \makecell[c]{HR, EDA, BVP,\\TEMP, PD,\\SA, and SV} & Valence and Arousal & Not Mentioned & \cmark & Young Adults\\
        
    \cmidrule{2-7}
        & \makecell[c]{Shu~\etal~\cite{shu2020}\\2020} & HR & \makecell[c]{Neutral, Happy and Sad} & \makecell[c]{Musics, Pictures\\and Videos} & \cmark & Young Adults\\
        
    \cmidrule{2-7}
        & \makecell[c]{Miranda~\etal~\cite{miranda2021}\\2021} & \makecell[c]{ECG, GSR,\\EEG and SKT} & Fear and Non-fear & Videos & \cmark & Adults\\
        
    \cmidrule{2-7}
        & \makecell[c]{Chen~\etal~\cite{chen2021}\\2021} & SP & \makecell[c]{Happiness, Sadness,\\Anger and Fear} & Videos & \xmark & Adults\\

    \cmidrule{2-7}
        & \makecell[c]{Badawi~\etal~\cite{badawi2024}\\2024} & \makecell[c]{PR, PRV, RR,\\ SCL} & \makecell[c]{Agitation\\and Aggression} & Indoor Activities & \cmark & Older Adults\\ 
    \midrule
    \midrule
    \multirow{15}{*}{\rotatebox[origin=c]{90}{Quantum}} &  \makecell[c]{Li~\etal~\cite{Li2021}\\2021} & \makecell[c]{Texts, Audio\\and Videos}& \makecell[c]{Happy, Sad, Neutral,\\ Angry, Excited, Frustrated} & \makecell[c]{Acts from TV} & \cmark & Young Adults\\

    \cmidrule{2-7}
        & \makecell[c]{Singh~\etal~\cite{singh2022}\\2022} & Videos & \makecell[c]{Anger, Disgust, Fear,\\Joy, Sadness, and Surprise} & \makecell[c]{Posed} & \cmark & Young Adults\\
        
    \cmidrule{2-7}
        & \makecell[c]{Li~\etal~\cite{li2023}\\2023} & \makecell[c]{Texts, Audio\\and Videos} & \makecell[c]{Neutral, Happy and Sad} & \makecell[c]{Speech and Interaction} & \cmark & Young Adults\\

    \cmidrule{2-7}
        & \makecell[c]{Mai~\etal~\cite{Mai2024}\\2024} & EEG & \makecell[c]{Anger, Disgust, Fear,\\Happiness, Sadness and Surprise} & \makecell[c]{Music Videos} & \xmark & Young Adults\\
        
    \cmidrule{2-7}
        & \makecell[c]{Liu~\etal~\cite{Liu2024}\\2024} & \makecell[c]{Texts, Audio\\and Videos} & \makecell[c]{Happy, Sad, Neutral,\\ Angry, Excited, Surprise} & \makecell[c]{Acts from TV} & \cmark & Adults\\

    \cmidrule{2-7}
        & \makecell[c]{Golchha~\etal~\cite{Golchha2025}\\2025} & \makecell[c]{Texts, Audio\\and Videos} & \makecell[c]{Anger, Disgust, Happiness,\\Fear, Sadness, Surprise and Neutral} & Posed & \xmark & Not Specific\\

    \cmidrule{2-7}
        & \makecell[c]{Singh~\etal~\cite{SINGH2025}\\2025} & \makecell[c]{Texts, Audio\\and Videos} & \makecell[c]{Anger, Disgust, Fear,\\Happiness, Sadness and Surprise} & Indoor Activities & \cmark & Not Specific\\ 
    \cmidrule{2-7}
        & \makecell[c]{\bf Proposed Work\\ \bf 2025} & \makecell[c]{\bf EDA, BVP, IBI} & \makecell[c]{\bf Positive, Negative\\ \bf and Neutral} & \bf TSST Protocol & \xmark & \bf Older Adults\\
    \bottomrule

    \end{tabular}
    }
    \footnotesize{\\SA: Saccadic amplitude, SV: Saccadic velocity, SKT: Skin Temperature, RR: Respiration Rate, PR: Pulse Rate, ECG: Electrocardiogram,\\EEG: Electroencephalography, EDA: Electrodermal Activity, SCL: Skin Conductance Level, PD: Pupil Dilation, HR: Heart Rate, TMP: Temperature,\\SC: Step Counter, BVP: Blood Volume Pressure, IBI: Inter Beat Interval, SP: Skin Potential, PRV: Pulse Rate Variability}
    \label{tab:comp}
\end{table*}

Table~\ref{tab:comp} provides some of the recent works on emotion detection in both classical and quantum domain. It showcases various methodologies, detected emotions, and participant demographics. While these studies focus primarily on adults and young adults, in order to address older adults, Badawi~\etal~\cite{badawi2024} used physiological markers to identify agitation and aggression during indoor activities. Biometric signals like skin temperature (ST), blood volume pulse (BVP), heart rate variability (HRV), and electrodermal activity (EDA) are typically used to make detections. However, for increased accuracy, the majority of current works combine facial emotion recognition~\cite{din2024, badawi2024, saganowski2022} with physiological data, or they concentrate on identifying a small number of emotions rather than a wide variety of affective states~\cite{schmidt2018}. Prior research frequently relies on cloud-based processing and deep learning models, which severely restricts the viability of real-time edge deployment~\cite{wijasena2021, ba2023}. This is concerning, especially for older persons who may feel uncomfortable or unwilling to participate in such assessments. The practical application of such a framework in privacy-sensitive settings, such as nursing homes or veterans' care facilities, is typically underexplored~\cite{rhodus2024}.

In parallel, researchers have moved toward using actual quantum circuits for emotion recognition tasks~\cite{Mai2024, singh2022, chandanwala2024}. Their studies employ Variational Quantum Circuits (VQCs) to process data from various sources such as EEG signals and facial expressions. These hybrid quantum-classical models are designed to extract meaningful features and perform classification that apply quantum computing principles to wearable and real-time emotional monitoring systems. Additionally, researchers further broaden the scope of QML in emotion detection~\cite{Golchha2025, SINGH2025, krishna2024, Alsubai2024} using Quantum inspired Neural Networks. They integrated quantum-enhanced processing with classical convolutional neural networks for facial emotion recognition. These efforts reflect a collective move towards harnessing the strengths of both quantum and classical paradigms to enhance emotion detection capabilities across various data types and application scenarios.

In response to the challenges posed by conventional emotion recognition methods, this work introduces a non-invasive framework that utilizes only physiological signals collected through wearable sensors removing the reliance on facial imagery or camera-based inputs. Our approach is specifically tailored for older adults, a population frequently underrepresented in emotion detection research despite being at increased risk for emotional and cognitive health issues. Unlike prior works that incorporate video or facial recognition technologies, our method is grounded solely in physiological data, offering a privacy-preserving and unobtrusive alternative. The system is designed for deployment in real-time inference in cloud that facilitates low-power, continuous monitoring in practical contexts such as nursing homes, healthcare facilities, and residential settings. By incorporating Quantum Machine Learning (QML) techniques, the proposed framework aims to effectively handle the high dimensionality and temporal complexity of physiological time-series data and achieve computational efficiency. The main contributions of this study are outlined as follows:

\begin{itemize}

    \item We used a dataset of 40 older adults whose emotions were recorded using facial expression and physiological signals from the Empatica E4 and Shimmer3 GSR+ wristbands.

    \item Live video was used to capture the participants' facial expressions, which were then labeled using iMotion's Facial Expression Analysis (FEA) module.

    \item Using solely labeled sensor data, we detected the intensity of three distinct emotions: neutral, positive, and negative. We showed that emotion detection can be accurately performed using simply sensor data, without the need for graphical or cognitive inputs.
    
    \item We performed classification tasks using both classical machine learning and hybrid quantum machine learning, and we provided our findings on the test set for different number of samples.

    \item We also showed that incorporating quantum machine learning can improve the overall performance reducing the false positive detection.
    
\end{itemize}

The paper is structured as follows:  Section \ref{sec_prot} describes the stimulus protocol for emotion and data recording.  The proposed method for emotion detection is provided in Section \ref{proposed_method}.  Section \ref{sec_results} summarizes the study's findings and analyses.  Section \ref{sec_conc} presents the conclusion and future prospects.

\section{Stimulus Protocol TSST and Data Recording}
\label{sec_prot}

This investigation involved a cohort of 40 healthy older adults, ranging in age from 60 to 80 years, with a demographic distribution of 28 females and 12 males. Following the exclusion of one participant's data due to corruption, the final dataset comprised 39 individuals. Before enrollment, participants underwent a screening process to exclude any pre-existing medical conditions and ensure the homogeneity of the sample. Emotional stimulation was achieved by applying the \textit{Trier Social Stress Test (TSST)} protocol.

\begin{figure}[htbp]
    \centering
    \includegraphics[width=\columnwidth]{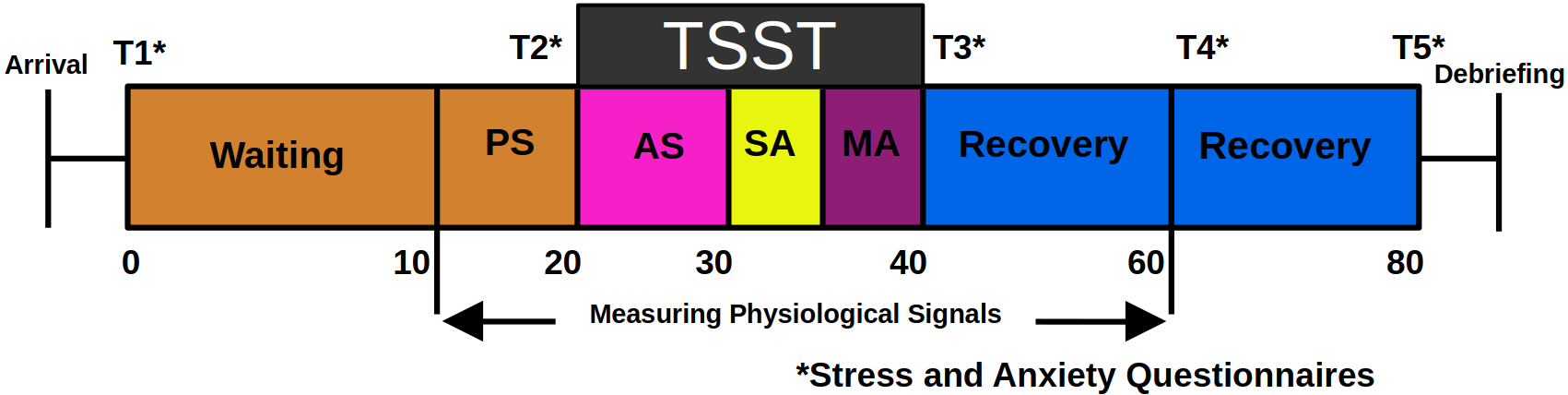}
    \caption{Stimulus Protocol TSST for Capturing Facial Emotion}
    \label{fig:exp_protocol}
\end{figure}

The TSST is a well-established experimental framework recognized for its efficacy in inducing psychological stress within a controlled yet realistic context~\cite{Birkett2011}. It was thus selected as the external emotional stimulus for this study. The experimental procedure, detailed in Figure~\ref{fig:exp_protocol}, consisted of sequential phases: a waiting period, a pre-stress baseline phase, a stress induction phase, and a recovery phase. During the initial waiting period, participants completed demographic questionnaires and were provided with an opportunity to clarify any procedural inquiries. Baseline physiological data were afterward acquired during the 20-minute pre-stress phase (T1-T2), immediately following the waiting period. The next 20-minute stress phase (T2-T3) incorporated a 10-minute anticipatory stress (AS) period, during which participants were instructed to prepare and deliver a continuous 5-minute speech before an audience. This was followed by a 5-minute segment (M) involving both a public speaking task and serial mental arithmetic. The arithmetic tasks involved basic addition and subtraction, with the difficulty escalating upon correct responses. The experimental session concluded with two 20-minute recovery phases (T3-T5).

Physiological sensor data were recorded using two commercially available wearable devices: the Empatica E4 and the Shimmer3 GSR+. These devices are equipped with integrated sensors for electrodermal activity (EDA), photoplethysmography (PPG), and skin temperature (ST). The continuous recording mode of these devices allowed for subsequent synchronization of the collected data following each experimental session. The measured physiological features included tri-axial accelerometer data (Yaw, Pitch, and Roll), temperature, internal analog-to-digital converter (ADC) voltage, galvanic skin response (GSR) resistance, heart rate, and GSR conductance.

%%%%%%%%%%%%%%%%%%%%%%%%%%%%%%%%%%%%%%%%%%%%%%%%%%%%%%%%%%%%%%%%%%%%%%%%%%%%%%%%
%                       Proposed Method for Emotion Detection
%%%%%%%%%%%%%%%%%%%%%%%%%%%%%%%%%%%%%%%%%%%%%%%%%%%%%%%%%%%%%%%%%%%%%%%%%%%%%%%%

\section{Proposed Method for Emotion Detection}
\label{proposed_method}

\begin{figure*}
    \centering
    \includegraphics[width=1.8\columnwidth]{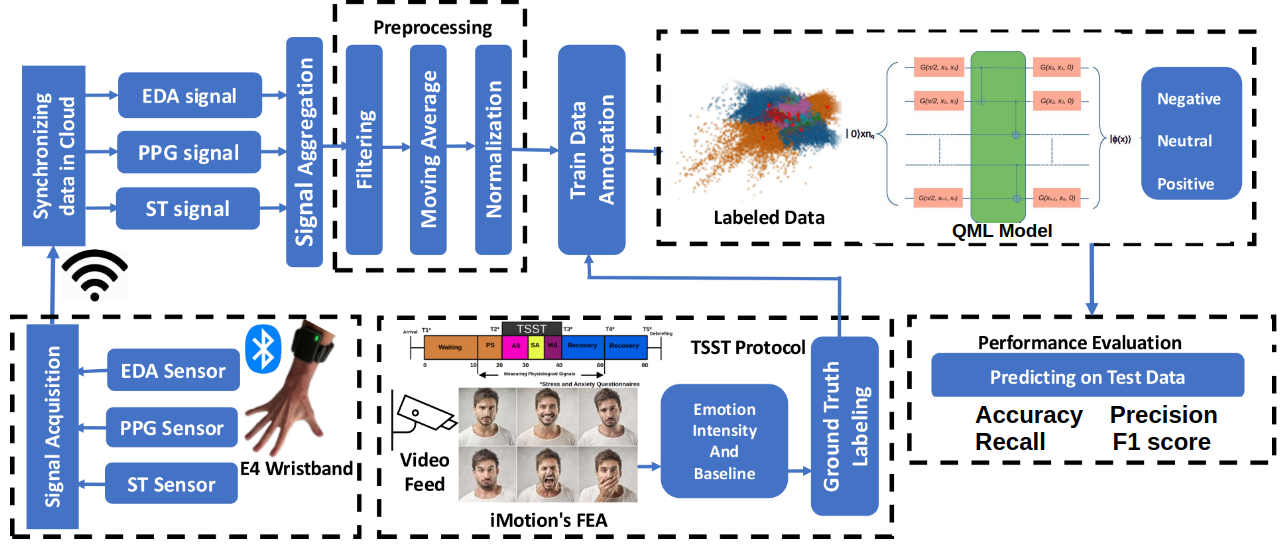}
    \caption{Proposed Emotion Detection Method}
    \label{fig:method}
\end{figure*}

In this section, we will discuss our proposed method for emotion detection with machine learning-based regression models. An overview of the proposed method is shown in Fig.~\ref{fig:method}. The three main components are (i) Data Preprocessing, (ii) Emotion ground truth labeling with iMotion's Facial Expression Analysis (FEA), and (iii) multiclass classification with classical and hybrid quantum machine learning models. We will now explain them in detail.

\subsection{Preprocessing and Ground Truth Labeling}
\label{gt}
After data collection, the dataset is preprocessed for labeling. During the preprocessing phase, the time series data was scanned using a 60-sample moving average window. This helped to eliminate undesirable artifacts and noise. To avoid data leakage, the train and test sets of participants are separated and normalized independently using z-score normalization. In our study, we create ground truth labeling for emotion recognition using iMotion's Facial Expression Analysis (FEA) module~\cite{imotions_fea_2024}, which leverages an enhanced implementation of the Facial Action Coding System (FACS). This is a popular framework in affective computing that can detect face muscle movements known as Action Units (AUs). Each combination of these action units can equate to a specific emotional expression. The FEA module uses computer vision and deep learning methods to assess these AUs in real time. These facial expressions were not used during training rather the sensor data and derived labels were used. This reinforces the privacy-preserving angle.

\begin{table}[htbp]
    \centering
    \caption{Emotion Baseline and Percentage Outside 1st Standard Deviation for Labeled Dataset}
    \renewcommand{\arraystretch}{1.3}
    \setlength{\tabcolsep}{2pt}
    \resizebox{\columnwidth}{!}{%
    \begin{tabular}{lccc||lcc}
        \hline
        \bf Emotion & \bf Baseline & \makecell[c]{\bf \% outside\\ \bf 1st std} & & \bf Emotion & \bf Baseline & \makecell[c]{\bf \% outside\\ \bf 1st std}\\
        \hline
        Joy & -0.6629763 & 24.28 & & Sadness & -0.515366 & 27.38 \\
        Anger & 0.3638698 & 41.11 & & \bf Neutral & -0.0408471 & 34.96 \\
        Surprise & -1.859378 & 7.61 & & \bf Positive & -0.6629763 & 24.28 \\
        Fear & -0.8419589 & 15.12 & & \bf Negative & 0.3638698 & 18.49 \\
        Contempt & -0.1459146 & 33.83 & & Confusion & 0.8094460 & 46.93 \\
        Disgust & -0.3184329 & 21.58 & & Frustration & 0.5876518 & 42.75 \\
        \hline
    \end{tabular}
    }
    \footnotesize{.\\$\star$ In This work, we will detect only \textit{Neutral, Positive} and \textit{Negative} emotion}
    \label{tab:dataset}
\end{table} \vspace{-0.1in}

This automated analysis ensures a uniform, objective way for identifying emotions that can avoid the inherent bias associated with self-report. We developed a robust dataset by synchronizing physiological sensor data (EDA, BVP, and IBI) with emotion labels extracted from facial expressions. This allows us to train and validate our framework. After labeling, Table~\ref{tab:dataset} displays the distribution of Emotion intensity. For that specific emotion class, samples outside of the first standard deviation are indicated as 0, while those inside are marked as 1. In this work, we plan to detect only 3 sets of emotions: \textit{Positive, Negative} and \textit{Neutral} as a part of multiclass classification problem where one emotion are assumed to be triggered separately from others.

\subsection{Hybrid Quantum Machine Learning Model}
\label{ml}

To assess the performance of our framework, we used seven traditional machine learning approaches to compare against proposed hybrid quantum machine learning model. They are Random Forest, K-Nearest Neighbors, Decision Tree, Gradient Boosting, Naive Bayes, Support Vector Machine and Logistic Regression. As a base hybrid model we use support vector machine with quantum kernel.

\begin{figure}[htbp]
    \centering
    \includegraphics[width=\columnwidth]{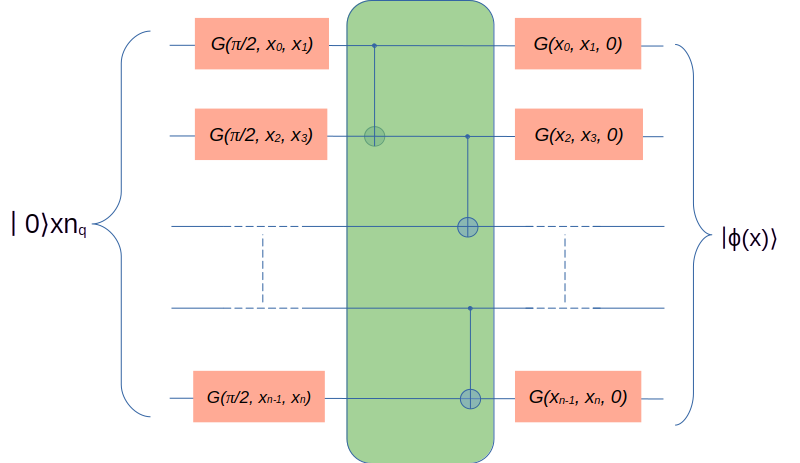}
    \caption{Data encoding circuit $U(x)$, for a data point $x$, that implements the feature map of the kernel. Here $G(\theta, \phi, \lambda)$ are 1 qubit gates and the entanglement gates correspond to CNOT gates}
    \label{fig:belis_kernel}
\end{figure}

\subsubsection{Data Encoding}
We used a quantum feature map that encodes two classical features per qubit initially proposed by Belis~\etal~\cite{Belis2024}. This maximizes information density while minimizing circuit depth. For each input vector $x$, the quantum circuit $U(x)$ applies a layer of parameterized $SU(2)$ rotation gates to each qubit. The rotation axes are not repeated within the same layer. After the rotation layer, nearest-neighbor CNOT gates are applied to introduce entanglement between adjacent qubits. This sequence of rotation and entanglement gates can be repeated to construct a deeper feature map. The final quantum state $ |\phi(x)\rangle = U(x)|0\rangle $ represents the encoded data point in a high-dimensional Hilbert space.

\subsubsection{Fidelity-Based Quantum Kernel}  
The similarity between two data points $x_i$ and $x_j$ is defined with the fidelity between their corresponding quantum states $k(x_i, x_j)$ defined in Equation~\eqref{eq:kernel}.
\begin{equation}
    k(x_i, x_j) = |\langle \phi(x_i) | \phi(x_j) \rangle|^2 = |\langle 0 | U^\dagger(x_i) U(x_j) | 0 \rangle|^2
    \label{eq:kernel}
\end{equation}

This expression computes the squared inner product between quantum states obtained by applying the parameterized circuits $U(x_i)$ and $U(x_j)$ to the all-zero initial state. As the fidelity is always between $0$ and $1$, the resulting Gram matrix $K$ is symmetric and a valid kernel function. The kernel maps the data to a non-linear quantum feature space that allows  the decision boundary to separate distant data points.

\subsubsection{Model Training and Testing}
\label{subsec_train}
As we had participant-specific data, there were multiple ways to generalize the machine learning models based on the dataset. To avoid data leakage, we split the dataset based on participants before normalizing it. We performed 5-fold cross-validation, keeping 20\% for testing and 80\% for training. For each of the models, the set of hyperparameters used is shown in Table~\ref{tab:hyp}. The values for these parameters were chosen based on the best-performing set on test data after multiple trials and errors.

\begin{table}[h]
    \centering
    \caption{Hyper-parameters Used to Train Machine Learning Models}
    \renewcommand{\arraystretch}{1}
    \setlength{\tabcolsep}{8pt}
    \resizebox{\columnwidth}{!}{%
    \begin{tabular}{ll}
        \toprule
        \bf Machine Learning Model & \bf Hyper-parameters\\
        
        \midrule
        Random Forest & \makecell[l]{no of estimators=100,\\criterion=squared error}\\
        
        \midrule
        Decision Tree & \makecell[l]{criterion=squared error}\\

        \midrule
        K-Nearest Neighbors (KNN) & \makecell[l]{no of neighbors=3, weights=uniform,\\distance metrics=minkowski}\\

        \midrule
        Gradient Boosting & \makecell[l]{no of estimators=100, Learning rate=0.1,\\loss=log\_loss, max depth=3}\\

        \midrule
        Multi Layer Perceptron & \makecell[l]{no of hidden layers=10, loss=mse,\\output layer activation=ReLU,\\optimizer=Adam, Learning rate=0.001,\\ maximum iteration=500}\\

        \midrule
        Support Vector Machine & \makecell[l]{Kernel=rbf, tolerance=$10^{-4}$}\\

        \midrule
        Logistic Regression & \makecell[l]{solver=lbfgs}\\

        \midrule
        \bf Quantum SVM & \makecell[l]{\bf Kernel=quantum\_kernel,\\ \bf tolerance=$10^{-4}$,
        n\_qubits=4}\\
        
        \bottomrule
    \end{tabular}
    \label{tab:hyp}
    }
\end{table}

Next, the quantum hybrid support vector machine was trained with the kernel calculated from the training and test sets. Given the quantum kernel matrix $K$, an SVM is used to perform the final classification. The optimization problem is formed as Equation~\eqref{svm} when $0 \leq \alpha_i \leq C, \quad \sum_{i=1}^{N} \alpha_i y_i = 0$.

\begin{equation}
 \max_{alpha} \sum_{i=1}^{N} \alpha_i - \frac{1}{2} \sum_{i,j=1}^{N} \alpha_i \alpha_j y_i y_j k(x_i, x_j)
 \label{svm}
\end{equation}

where $\alpha_i$ are the coefficients, $y_i \in \{-1, 1\}$ are the class labels, and $C$) is a regularization parameter. During inference, the decision function for a new input $x$ is given by
$f(x) = \text{sign} \left( \sum_{i=1}^{N} \alpha_i y_i k(x_i, x) + b \right)$, where $b$ is the bias term. 

We utilized \textit{python 3.10} and the \textit{Qiskit} library to compute quantum kernel matrices for emotion detection. The quantum kernel is constructed using the \textit{FidelityQuantumKernel} from \textit{Qiskit}'s \textit{machine\_learning} module that computes the inner products of these quantum states. The simulation is executed on a classical backend with \textit{AerSimulator} using the \textit{statevector} method to compute the fidelity between states. The resulting training and test kernel matrices are forwarded to SVM and optimized with \textit{SVC} package of the \textit{scikit-learn} module.

%%%%%%%%%%%%%%%%%%%%%%%%%%%%%%%%%%%%%%%%%%%%%%%%%%%%%%%%%%%%%%%%%%%%%%%%%%%%%%%%
%                       Result Analysis and Discussion
%%%%%%%%%%%%%%%%%%%%%%%%%%%%%%%%%%%%%%%%%%%%%%%%%%%%%%%%%%%%%%%%%%%%%%%%%%%%%%%%

\section{Result Analysis and Discussion}
\label{sec_results}
In this section, we will evaluate the performance of machine learning models with our dataset with respect to the estimated ground truth mentioned in Section~\ref{gt}. 

\begin{table*}[htbp]
    \centering
    \caption{Performance of Classical ML Models for 2000 Samples (avg of 5 folds)}    
    \renewcommand{\arraystretch}{1.25}
    \setlength{\tabcolsep}{3pt}
    \resizebox{2\columnwidth}{!}{%
        \begin{tabular}{l|cccc||cccc||cccc}
        \hline
        \multirow{2}{*}{\bf ML Algorithm} & \multicolumn{4}{c||}{\bf Negative} & \multicolumn{4}{c||}{\bf Neutral} & \multicolumn{4}{c}{\bf Positive}\\

        \cline{2-13}
         & \bf{Accuracy} & \bf{Precision} & \bf{Recall} & \bf{F1 Score} & \bf{Accuracy} & \bf{Precision} & \bf{Recall} & \bf{F1 Score} & \bf{Accuracy} & \bf{Precision} & \bf{Recall} & \bf{F1 Score} \\
         
        \hline
        Random Forest & 0.796 & 0.768 & 0.619 & 0.685 & 0.901 & 0.925 & 0.788 & 0.851 & 0.885 &	0.853 &	0.688 &	0.761\\
        
        \hline
        K-Nearest Neighbors & 0.762 & 0.662 & 0.600 & 0.629 & 0.863 & 0.750 & 0.680 & 0.713 & 0.846 & 0.735	& 0.667	& 0.699\\
        
        \hline
        Decision Tree & 0.747 & 0.614 & 0.615 & 0.615 & 0.846 & 0.696 &	0.697 &	0.697 & 0.830 &	0.682 &	0.684 &	0.683\\
        
        \hline
        Gradient Boosting & 0.699 & 0.710 & 0.203 & 0.316 & 0.792 &	0.804 &	0.230 &	0.358 &	0.776 &	0.789 &	0.226 &	0.351\\
        
        \hline
        Naive Bayes & 0.655 & 0.434 & 0.201 & 0.274 & 0.742 &	0.492 &	0.227 &	0.311 &	0.728 &	0.482 &	0.223 &	0.305\\
        
        \hline
        Support Vector Machine & 0.685 & 0.720 & 0.132 & 0.223 & 0.777 &	0.816 &	0.149 &	0.253 &	0.761 &	0.800 &	0.146 &	0.248\\ 
        
        \hline
        Logistic Regression & 0.664 & 0.635 & 0.036 & 0.068 & 0.753 & 0.720 & 0.041	& 0.077	& 0.738	& 0.705	& 0.040	& 0.075\\
        
        \hline
    \end{tabular}

    \label{tab:classical}
    }
\end{table*}

\begin{table*}[!h]
    \centering
    \caption{Performance of Hybrid Quantum SVM Model with Belis Kernel (avg of 5 folds)}
    \renewcommand{\arraystretch}{1.25}
    \setlength{\tabcolsep}{4pt}
    \resizebox{2\columnwidth}{!}{%
        \begin{tabular}{l|cccc||cccc||cccc}
        \hline
        \multirow{2}{*}{\bf No of Samples} & \multicolumn{4}{c||}{\bf Negative} & \multicolumn{4}{c||}{\bf Neutral} & \multicolumn{4}{c}{\bf Positive}\\

        \cline{2-13}
         & \bf{Accuracy} & \bf{Precision} & \bf{Recall} & \bf{F1 Score} & \bf{Accuracy} & \bf{Precision} & \bf{Recall} & \bf{F1 Score} & \bf{Accuracy} & \bf{Precision} & \bf{Recall} & \bf{F1 Score} \\
         
        \hline
        train: 640; test: 160 & 0.800 &	0.853 &	0.860 &	0.851 & 0.863 &	0.751 &	0.803 &	0.775 & 0.892 &	0.878 &	0.938 &	0.905\\
        
        \hline
        train: 1600; test: 400 & 0.815 & 0.887 & 0.864 & 0.872 & 0.874 &	0.790 &	0.864 &	0.824 & 0.881 &	0.896 &	0.898 & 0.897\\
        
        \hline
    \end{tabular}

    \label{tab:quantum}
    }
\end{table*}

The results presented in Table~\ref{tab:classical} provide a comparative analysis of traditional machine learning algorithms for the emotion classification task with 2,000 samples. Random Forest stands out as the top-performing model by consistently achieving the highest F1 scores across the Negative (0.685), Neutral (0.851), and Positive (0.761) classes. These results highlight its strong balance between precision and recall. Tree-based approaches, such as Random Forest and Decision Tree, demonstrate superior performance over linear models, particularly for the Neutral and Positive classes. This suggests that tree-based models work better to capture the non-linear relationships inherent in emotion data. In contrast, Logistic Regression and Support Vector Machine (SVM) with rbf kernel exhibit relatively high precision but notably low recall. This imbalance leads to reduced F1 scores.

On the other hand, Table~\ref{tab:quantum} shows that hybrid quantum SVMs utilizing the Belis kernel exhibit clear performance advantages. The quantum model surpasses classical models in most classes, even with smaller samples. With only 640 training samples, the quantum SVM achieves an F1 score of 0.905 for the Positive class that exceeds the best classical result of 0.761 from Random Forest. When the training set increases to 1,600 samples, the hybrid quantum model’s robustness and generalization further improve. The 36\% increase in recall value shows that the model maintains strong precision-recall balance. This proves our method to be reliable without overfitting or excessive false positives. These results suggest that quantum kernels can capture complex data correlations that classical kernels may overlook.

%%%%%%%%%%%%%%%%%%%%%%%%%%%%%%%%%%%%%%%%%%%%%%%%%%%%%%%%%%%%%%%%%%%%%%%%%%%%%%%%
%                       Conclusion
%%%%%%%%%%%%%%%%%%%%%%%%%%%%%%%%%%%%%%%%%%%%%%%%%%%%%%%%%%%%%%%%%%%%%%%%%%%%%%%%

\section{Conclusion}
\label{sec_conc}

This work establishes the viability of detecting emotional states exclusively through physiological signals, avoiding facial recognition technologies. The findings indicate that while classical machine learning models achieve satisfactory results in estimating emotion intensities, the adoption of a hybrid quantum machine learning (QML) model yields notable improvements in both accuracy and robustness. Our experiments with the quantum kernel based SVM consistently outperformed classical models across all emotion classes, even with smaller samples. These outcomes underscore the effectiveness of combining wearable sensor data with quantum-enhanced algorithms for emotion recognition while maintaining user privacy. The implications of this work are particularly relevant for populations such as individuals with Alzheimer’s Disease and Related Dementias (ADRD), veterans experiencing Post-Traumatic Stress Disorder (PTSD), and others with cognitive or emotional difficulties. The proposed methodology can be readily implemented in clinical and assisted living environments to enable passive, unobtrusive monitoring of emotional well-being. Future research should explore more advanced QML architectures like Quantum Neural Networks and investigate real-time deployment on edge computing platforms.

\section{Acknowledgments}
This research used resources of the Oak Ridge Leadership Computing Facility, which is a DOE Office of Science User Facility supported under Contract DE-AC05-00OR22725.



\bibliographystyle{style}
\bibliography{reference} \balance

\begin{thebibliography}{10}
\providecommand{\url}[1]{#1}
\csname url@samestyle\endcsname
\providecommand{\newblock}{\relax}
\providecommand{\bibinfo}[2]{#2}
\providecommand{\BIBentrySTDinterwordspacing}{\spaceskip=0pt\relax}
\providecommand{\BIBentryALTinterwordstretchfactor}{4}
\providecommand{\BIBentryALTinterwordspacing}{\spaceskip=\fontdimen2\font plus
\BIBentryALTinterwordstretchfactor\fontdimen3\font minus
  \fontdimen4\font\relax}
\providecommand{\BIBforeignlanguage}[2]{{%
\expandafter\ifx\csname l@#1\endcsname\relax
\typeout{** WARNING: IEEEtran.bst: No hyphenation pattern has been}%
\typeout{** loaded for the language `#1'. Using the pattern for}%
\typeout{** the default language instead.}%
\else
\language=\csname l@#1\endcsname
\fi
#2}}
\providecommand{\BIBdecl}{\relax}
\BIBdecl

\bibitem{rhodus2024}
E.~K. Rhodus, M.~S.~H. Onim, C.~Roberts, S.~Kumar, A.~M. Burhan, and
  H.~Thapliyal, ``Utilization of wearable devices as means for remote digital
  biometric data collection in a community-based, rural randomized controlled
  trial among alzheimer’s disease dyads,'' \emph{Alzheimer's \& Dementia},
  vol.~20, p. e090883, 2024.

\bibitem{din2024}
I.~U. Din, A.~Almogren, J.~J. Rodrigues, and A.~Altameem, ``Advancing secure
  and privacy-preserved decision-making in iot-enabled consumer electronics via
  multimodal data fusion,'' \emph{IEEE Transactions on Consumer Electronics},
  2024.

\bibitem{badawi2024}
A.~Badawi, S.~Elmoghazy, S.~Choudhury, S.~Elgazzar, K.~Elgazzar, and A.~Burhan,
  ``A novel multimodal system to predict agitation in people with dementia
  within clinical settings: A proof of concept,'' \emph{arXiv preprint
  arXiv:2411.08882}, 2024.

\bibitem{onim2023review}
M.~S.~H. Onim, E.~Rhodus, and H.~Thapliyal, ``A review of context-aware machine
  learning for stress detection,'' \emph{IEEE Consumer Electronics Magazine},
  2023.

\bibitem{Delmastro2020}
F.~Delmastro, F.~Di~Martino, and C.~Dolciotti, ``Cognitive training and stress
  detection in mci frail older people through wearable sensors and machine
  learning,'' \emph{IEEE Access}, vol.~8, pp. 65\,573--65\,590, 2020.

\bibitem{Kikhia2016}
B.~Kikhia, T.~G. Stavropoulos, S.~Andreadis, N.~Karvonen, I.~Kompatsiaris,
  S.~S{\"a}venstedt, M.~Pijl, and C.~Melander, ``Utilizing a wristband sensor
  to measure the stress level for people with dementia,'' \emph{Sensors},
  vol.~16, no.~12, p. 1989, 2016.

\bibitem{onim2024utilizing}
M.~S.~H. Onim, H.~Thapliyal, and E.~K. Rhodus, ``Utilizing machine learning for
  context-aware digital biomarker of stress in older adults,''
  \emph{Information}, vol.~15, no.~5, p. 274, 2024.

\bibitem{Cheong2020}
S.-M. Cheong, C.~Bautista, and L.~Ortiz, ``Sensing physiological change and
  mental stress in older adults from hot weather,'' \emph{IEEE Access}, vol.~8,
  pp. 70\,171--70\,181, 2020.

\bibitem{Ferreira2014}
E.~Ferreira, D.~Ferreira, S.~Kim, P.~Siirtola, J.~R{\"o}ning, J.~F. Forlizzi,
  and A.~K. Dey, ``Assessing real-time cognitive load based on
  psycho-physiological measures for younger and older adults,'' in \emph{2014
  IEEE Symp. on Computational Intelligence, Cognitive Algorithms, Mind, and
  Brain}, 2014, pp. 39--48.

\bibitem{zhang2020}
T.~Zhang, A.~El~Ali, C.~Wang, A.~Hanjalic, and P.~Cesar, ``Corrnet:
  Fine-grained emotion recognition for video watching using wearable
  physiological sensors,'' \emph{Sensors}, vol.~21, no.~1, p.~52, 2020.

\bibitem{shu2020}
L.~Shu, Y.~Yu, W.~Chen, H.~Hua, Q.~Li, J.~Jin, and X.~Xu, ``Wearable emotion
  recognition using heart rate data from a smart bracelet,'' \emph{Sensors},
  vol.~20, no.~3, p. 718, 2020.

\bibitem{miranda2021}
J.~A. Miranda, M.~F.~Canabal, L.~Gutierrez-Martin, J.~M. Lanza-Gutierrez,
  M.~Portela-Garcia, and C.~Lopez-Ongil, ``Fear recognition for women using a
  reduced set of physiological signals,'' \emph{Sensors}, vol.~21, no.~5, p.
  1587, 2021.

\bibitem{chen2021}
S.~Chen, K.~Jiang, H.~Hu, H.~Kuang, J.~Yang, J.~Luo, X.~Chen, and Y.~Li,
  ``Emotion recognition based on skin potential signals with a portable
  wireless device,'' \emph{Sensors}, vol.~21, no.~3, p. 1018, 2021.

\bibitem{Li2021}
Q.~Li, D.~Gkoumas, A.~Sordoni, J.-Y. Nie, and M.~Melucci, ``Quantum-inspired
  neural network for conversational emotion recognition,'' \emph{Proceedings of
  the AAAI Conference on Artificial Intelligence}, vol.~35, no.~15, pp.
  13\,270--13\,278, May 2021.

\bibitem{singh2022}
J.~Singh, F.~Ali, B.~Shah, K.~S. Bhangu, and D.~Kwak, ``Emotion quantification
  using variational quantum state fidelity estimation,'' \emph{IEEE Access},
  vol.~10, pp. 115\,108--115\,119, 2022.

\bibitem{li2023}
Z.~Li, Y.~Zhou, Y.~Liu, F.~Zhu, C.~Yang, and S.~Hu, ``{QAP}: A quantum-inspired
  adaptive-priority-learning model for multimodal emotion recognition,'' in
  \emph{Findings of the Association for Computational Linguistics: ACL 2023},
  A.~Rogers, J.~Boyd-Graber, and N.~Okazaki, Eds.\hskip 1em plus 0.5em minus
  0.4em\relax Toronto, Canada: Association for Computational Linguistics, Jul.
  2023, pp. 12\,191--12\,204.

\bibitem{Mai2024}
N.-D. Mai, H.~Barki, and W.-Y. Chung, ``A wearable bte-eeg embedded device for
  emotion monitoring with quantum machine learning,'' in \emph{2024 Tenth
  International Conference on Communications and Electronics (ICCE)}, 2024, pp.
  562--566.

\bibitem{Liu2024}
Y.~Liu, Y.~Zhang, and D.~Song, ``A quantum probability driven framework for
  joint multi-modal sarcasm, sentiment and emotion analysis,'' \emph{IEEE
  Transactions on Affective Computing}, vol.~15, no.~1, pp. 326--341, 2024.

\bibitem{Golchha2025}
R.~Golchha, M.~Sahu, and V.~Bhateja, ``Quantum-based deep learning method for
  recognition of facial expressions,'' \emph{Neural Computing and
  Applications}, mar 2025.

\bibitem{SINGH2025}
J.~Singh, K.~S. Bhangu, A.~Alkhanifer, A.~A. AlZubi, and F.~Ali, ``Quantum
  neural networks for multimodal sentiment, emotion, and sarcasm analysis,''
  \emph{Alexandria Engineering Journal}, vol. 124, pp. 170--187, 2025.

\bibitem{saganowski2022}
S.~Saganowski, J.~Komoszy{\'n}ska, M.~Behnke, B.~Perz, D.~Kunc, B.~Klich,
  {\L}.~D. Kaczmarek, and P.~Kazienko, ``Emognition dataset: emotion
  recognition with self-reports, facial expressions, and physiology using
  wearables,'' \emph{Scientific data}, vol.~9, no.~1, p. 158, 2022.

\bibitem{schmidt2018}
P.~Schmidt, A.~Reiss, R.~Duerichen, C.~Marberger, and K.~Van~Laerhoven,
  ``Introducing wesad, a multimodal dataset for wearable stress and affect
  detection,'' in \emph{Proceedings of the 20th ACM international conference on
  multimodal interaction}, 2018, pp. 400--408.

\bibitem{wijasena2021}
H.~Z. Wijasena, R.~Ferdiana, and S.~Wibirama, ``A survey of emotion recognition
  using physiological signal in wearable devices,'' in \emph{International
  Conference on Artificial Intelligence and Mechatronics Systems}.\hskip 1em
  plus 0.5em minus 0.4em\relax IEEE, 2021, pp. 1--6.

\bibitem{ba2023}
S.~Ba and X.~Hu, ``Measuring emotions in education using wearable devices: A
  systematic review,'' \emph{Computers \& Education}, vol. 200, p. 104797,
  2023.

\bibitem{chandanwala2024}
A.~A. Chandanwala, S.~Bhowmik, P.~Chaudhury, and S.~C. Pravin, ``Hybrid quantum
  deep learning model for emotion detection using raw eeg signal analysis,''
  2024.

\bibitem{krishna2024}
S.~Suraj Jai~Krishna, M.~Anish, A.~M. Posonia, J.~Albert~Mayan, and P.~Asha,
  ``Gesture and emotion detection using quantum computing for enhanced
  recognition and analysis,'' in \emph{2024 International Conference on Expert
  Clouds and Applications (ICOECA)}, 2024, pp. 530--535.

\bibitem{Alsubai2024}
S.~Alsubai, A.~Alqahtani, A.~Alanazi, M.~Sha, and A.~Gumaei, ``Facial emotion
  recognition using deep quantum and advanced transfer learning mechanism,''
  \emph{Frontiers in Computational Neuroscience}, vol. Volume 18 - 2024, 2024.

\bibitem{Birkett2011}
M.~A. Birkett, ``The trier social stress test protocol for inducing
  psychological stress,'' \emph{JoVE (Journal of Visualized Experiments)},
  no.~56, p. e3238, 2011.

\bibitem{imotions_fea_2024}
\BIBentryALTinterwordspacing
{iMotions A/S}, ``Facial expression analysis (fea),'' 2024, accessed:
  2025-03-22. [Online]. Available:
  \url{https://imotions.com/blog/publications/}
\BIBentrySTDinterwordspacing

\bibitem{Belis2024}
V.~Belis, K.~A. Wo{\'{z}}niak, and E.~Puljak, ``Quantum anomaly detection in
  the latent space of proton collision events at the lhc,''
  \emph{Communications Physics}, vol.~7, no.~1, p. 334, Oct 2024.

\end{thebibliography}

\end{document}